\documentclass[11pt]{article}

\usepackage[]{acl}
\usepackage{xcolor}
\usepackage{times}
\usepackage{latexsym}
\usepackage{placeins}

\usepackage[T1]{fontenc}

\usepackage[utf8]{inputenc}

\usepackage{microtype}

\usepackage{inconsolata}

\usepackage{graphicx}

\usepackage{amsmath}
\usepackage{hyperref}
\usepackage[nameinlink]{cleveref}
\usepackage{multirow}
\usepackage{arydshln}
\usepackage{subcaption}
\usepackage{lipsum}
\usepackage{xcolor}
\usepackage{CJKutf8}  
\usepackage{booktabs}
\usepackage{amsfonts}
\usepackage{amssymb}
\usepackage{pifont}
\usepackage{fdsymbol}

\newcommand{\cmark}{\ding{51}}%
\newcommand{\xmark}{\ding{55}}%

\newcommand{\framework}{\textsc{ConFit v2}}
\newcommand{\confitold}{\textsc{ConFit}}
\newcommand{\confitsimple}{\textsc{ConFit}$^{*}$}
\newcommand{\RunnerUpMining}{\textsc{Runner-Up Mining}}
\newcommand{\RunnerUpMiningShort}{\textsc{RUM}}
\newcommand{\HyRe}{\textsc{Hypothetical Resume Embedding}}

\newcommand{\HyReShort}{\textsc{HyRe}}

\newcommand{\chinese}[1]{\begin{CJK}{UTF8}{gbsn}#1\end{CJK}}

\title{\framework{}: Improving Resume-Job Matching using Hypothetical Resume Embedding and Runner-Up Hard-Negative Mining}

\author{Xiao Yu\thanks{denotes equal contribution}$^{\spadesuit}$
        Ruize Xu$^{*\spadesuit}$
        Chengyuan Xue$^{*\vardiamondsuit}$
        Jinzhong Zhang$^{\clubsuit}$\label{author:intellipro}
        Xu Ma$^{\clubsuit}$\label{author:intellipro}
        Zhou Yu$^\spadesuit$~~~\\[3pt]
  $^\spadesuit$Columbia University~ 
  $^\vardiamondsuit$University of Toronto~
  $^\clubsuit$Intellipro Group Inc.\\[3pt] 
  \texttt{\{xy2437,zy2461\}@columbia.edu}\\
  \texttt{\{jinzhong\}@intelliprogroup.com}
}

\begin{document}
\maketitle

\begin{abstract}
  A reliable resume-job matching system helps a company recommend suitable candidates from a pool of resumes and helps a job seeker find relevant jobs from a list of job posts.
  However, since job seekers apply only to a few jobs, interaction labels in resume-job datasets are sparse.
  We introduce \framework{}, an improvement over \confitold{} to tackle this sparsity problem.
  We propose two techniques to enhance the encoder’s contrastive training process: augmenting job data with hypothetical reference resume generated by a large language model; and creating high-quality hard negatives from unlabeled resume/job pairs using a novel hard-negative mining strategy.
 We evaluate \framework{} on two real-world datasets and demonstrate that it outperforms \confitold{} and prior methods (including BM25 and OpenAI text-embedding-003), achieving an average \emph{absolute} improvement of 13.8\% in recall and 17.5\% in nDCG across job-ranking and resume-ranking tasks.
 
\end{abstract}

\section{Introduction}
Online recruitment platforms like LinkedIn serve over 990 million users and 65 million businesses, processing more than 100 million job applications each month \cite{linkedin}. As these platforms continue to grow, there is a rising need for \emph{efficient} and \emph{robust} resume-job matching systems.
A practical system that reliably identifies suitable candidates/jobs from large pools will save considerable effort for both employers and job seekers.

Since both resume and job posts are often stored as text data, there has been an increased interest in using transformer models to model resume-job fit (or referred to as ``person-job fit'').
Many prior works \cite{pjfnn,APJFNN,mvcon,DPGNN,InEXIT} focus on designing complex modeling techniques to model resume-job matching.
For example, APJFNN \cite{APJFNN} uses hierarchical recurrent neural networks to process the job and resume content, and DPGNN \cite{DPGNN} uses a dual-perspective graph neural network to model the relationship between resumes and jobs.
Unlike these manual model/feature engineering methods, more recently \citet{confit_v1} shows that dense retrieval techniques like contrastive learning achieve significant improvements without relying on any complex designs.
By introducing dense retrieval methods to resume-job matching, \confitold{} \cite{confit_v1} presented a simple yet effective method to rank thousands of resumes/jobs in milliseconds, by combining training neural encoders such as E5 \cite{E5} with inner-product search algorithms such as FAISS \cite{FAISS}.

\begin{figure}[!t]
    \centering
    \includegraphics[scale=0.35]{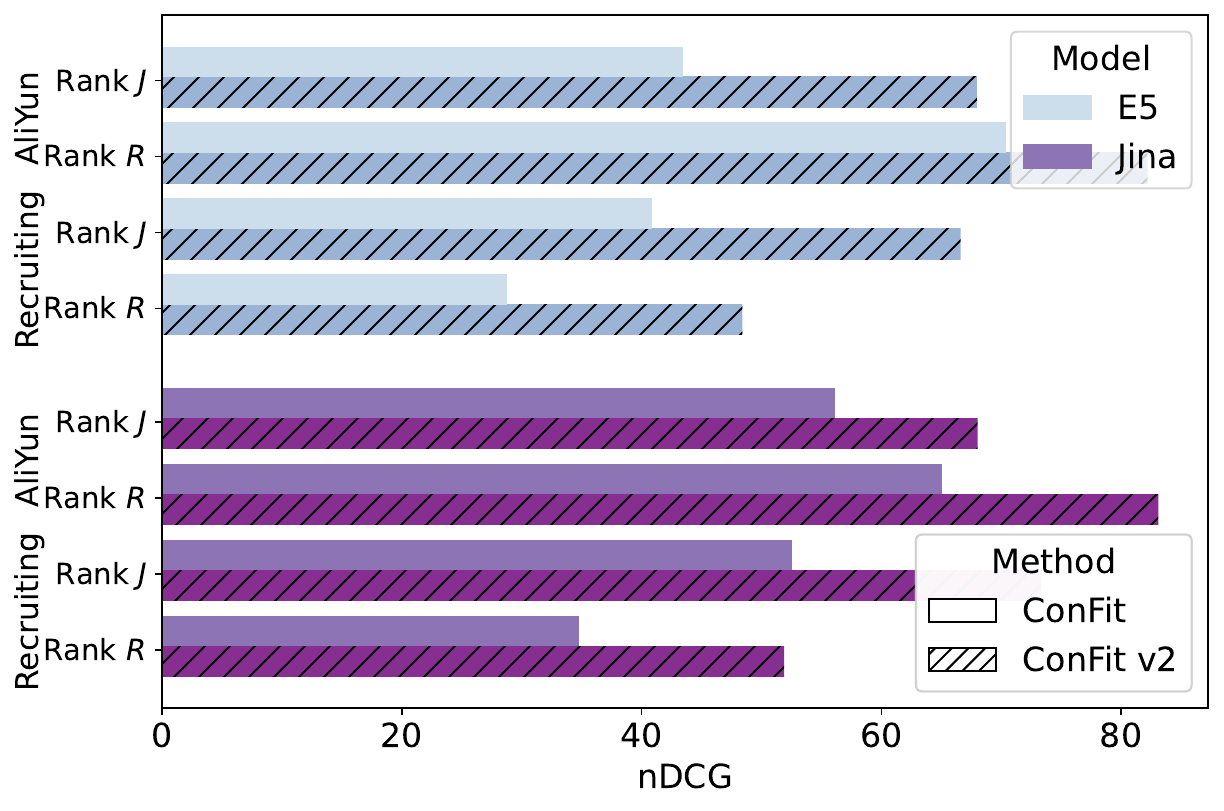}
    \caption{Performance comparison between \framework{} and \confitold{} across the AliYun and the Recruiting dataset. ``Rank $R$'' indicates ranking resume, and ``Rank $J$'' indicates ranking job.
    }
    \label{fig:confit_v1_vs_v2}
\end{figure}

\begin{figure*}[!t]
    \centering
    \includegraphics[scale=0.78]{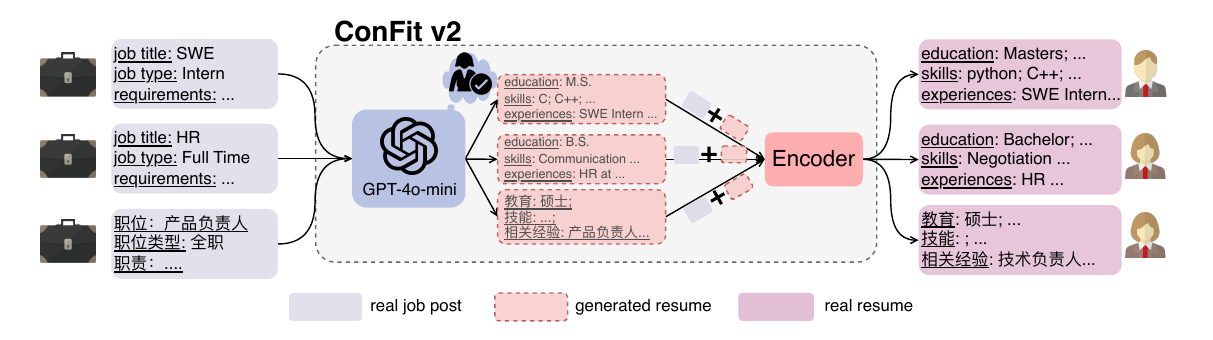}
    \caption{\framework{} inference. Given a job post, we first use an LLM to generate a hypothetical reference resume given the job post, and then outputs a job embedding using the concatenation of the generated resume and the job post. Given a resume, \framework{} outputs a resume embedding directly using our trained encoder model. Finally, cosine similarity is used to compute the compatibility between the input resume and job post.
    }
    \label{fig:inference}
\end{figure*}

We propose \framework{}, an enhanced baseline that achieves an average absolute improvement of 13.8\% in recall and 17.5\% in nDCG on ranking resumes and jobs compared to \confitold{} (see \Cref{fig:confit_v1_vs_v2}).
We introduce two key improvements to enhance the neural encoder's training process: 
1) \underline{Hy}pothetical \underline{R}esume \underline{E}mbedding (\HyReShort{}) that leverages a large language model (LLM) to generate a hypothetical resume and augment the job post (see \Cref{fig:inference}), providing implicit details to reduce the burden on encoder training; and
2) \underline{R}unner-\underline{U}p \underline{M}ining (\RunnerUpMiningShort{}) that mines a large set of high-quality hard negatives to help the encoder better discern positive with near-positive samples.
We train \framework{} using \HyReShort{} and \RunnerUpMiningShort{} on two real-world datasets, and show that \framework{} achieves new state-of-the-art performance on ranking resumes and jobs, outperforming \confitold{} and other prior work (including BM25 and OpenAI text-embedding-003-large).
We will open-source our code and data (under license agreements) to provide a strong baseline for future research in empowering resume-job matching systems with dense retrieval techniques.

\section{Background}
\label{sec:Background}
A resume-job matching (or often called \emph{person-job fit}) system models the compatibility between a resume $R$ and a job post $J$, allowing the systems to recommend the most suitable candidates for a given job, or suggest the most relevant jobs for a given candidate \cite{mvcon,DPGNN,InEXIT,confit_v1}.
Since resumes and job posts are often stored as text data, many prior works consider using neural networks (e.g., encoders) to quantify the compatibility between resumes and jobs:
\[
\textrm{score}(R, J) = f_\theta(R, J) \to \mathbb{R},
\] 
where $f_\theta$ could be directly modeled by a neural network \cite{pjfnn,DPGNN,InEXIT}, or by computing inner product/cosine similarity \cite{confit_v1} between a resume embedding $f_\theta(R)$ and a job embedding $f_\theta(J)$.

Despite the rapid growth of online recruitment platforms and the increasing availability of job posts and resume data, there are \emph{very few interaction labels} between any resumes and job posts \cite{mvcon,confit_v1}.
This is because a candidate usually applies to only a few positions, and a job interviewer often reviews only a few resumes for a given job post.
Often, the resulting dataset $\mathcal{D}= \{ R_i, J_i, y_i \}$ has size $|\mathcal{D}| \ll n_{R} \times n_{J}$, where $n_{R}$ and $n_{J}$ are the total number of resumes and jobs respectively, and $y_{i} \in \{0,1\}$ is a \emph{binary} signal representing whether a resume $R_i$ is short listed for a job $J_i$.
For example, in both the Recruiting and AliYun datasets (\Cref{subsec:Dataset and Preprocessing}) used in this work, \textbf{less than 0.05\%} of the total possible (resume, job) pairs are annotated.
This label sparsity poses challenges in: 1) crafting high-quality training data/hard negatives for training a neural encoder $f_\theta$; and 2) learning a representation space that generalizes well across diverse resumes and job posts.

\section{Approach}
\label{sec:ConFit}

We propose \framework{}, a simple and general-purpose approach that improves \confitold{} \cite{confit_v1} to model resume-job compatibility.
Similar to \confitold{}, we use an encoder to produce an embedding of a given resume or job post, and model the matching score between an $\langle R,J \rangle$ pair as the cosine similarity of their representations.
Unlike \confitold{}, \framework{} uses a simpler encoder architecture, and substantially improves ranking performance by using 1) hypothetical reference resume generated by an LLM to help improve the encoder's job data understanding; and 2) a novel hard-negative mining method to enhance encoder contrastive learning.
We provide a high-level overview of \framework{} during inference in \Cref{fig:inference}.
We detail each modification below.

\subsection{Encoder Architecture}
\label{subsec:Model Architecture}
Many prior works on resume-job matching employ complex neural architectures to encode human-designed matching heuristics \cite{pjfnn,DPGNN}.
For example, \citet{InEXIT,confit_v1} uses a hierarchical attention mechanism to model interactions between each \emph{section} (e.g., Education, Experience, etc.) of a resume and a job post.
However, in practice, these modifications often rely on specific domain knowledge and require additional structural constraints on resume/job data, restricting their generalizability.

Different from these methods, we treat the resume/job post as \emph{a single sequence of text}, and directly use a transformer-based encoder to produce the embedding of the entire document.
This allows the encoder model itself to learn the most relevant features from the data, instead of relying on human-designed heuristics.
We illustrate this difference in \Cref{fig:archi_diff}.
In practice, we find this simpler design is more effective and robust across different datasets and backbones (\Cref{subsec:Ablation Studies}).
We denote this simplified encoder as \confitsimple{}.



\subsection{\HyRe{}}
\label{subsec:Hypothetical Resume Embedding}
In traditional dense passage retrieval tasks, one challenge is the discrepancy between query and passage formats and content \cite{hyde,hyqe}.
To address this, \citet{hyde} shows that zero-shot passage ranking improves significantly with Hypothetical Document Embeddings (HyDe), where an LLM generates a hypothetical passage from a query, and the encoder ranks actual passages based on their cosine similarity to the hypothetical one.
We observe a similar discrepancy in resume-job matching, and find a substantial performance gain when high-quality (real) resumes are used in place of jobs during ranking. See \Cref{sec:Additional Results on HyRe} for more details. 

Given a dataset of accepted/rejected resume-job pairs, one extension to systems like HyDe is to finetune an LLM such as LLaMA-3 \cite{grattafiori2024llama3herdmodels} or Qwen-2.5 \cite{qwen2.5} using SFT and DPO \cite{rafailov2024directpreferenceoptimizationlanguage}. However, in our prior study, we find fine-tuning these models yields inferior results (see \Cref{subsec:Finetuning HyRe}) to few-shot prompting powerful closed-source models such as GPT-4o-mini \cite{4omini}.
We believe this is because accepting a resume can be subjective, making it hard for an LLM to learn an ``ideal candidate'' from the dataset.
We thus construct hypothetical references by \emph{augmenting the existing job posts}. Specifically, we 1) fix $N$ accepted resume-job pairs from the training set as few-shot examples; 2) prompt GPT-4o-mini to generate a reference resume given a job post; and 3) concatenate the generated resume with the real job post. We repeat this for all jobs used during training and testing.

\subsection{\RunnerUpMining{}}
\label{subsec:Hard Negative Mining}
To accurately model resume-job compatibility, the encoder needs to distinguish between matching and near-matching resume-job pairs.
\confitold{} \cite{confit_v1} achieves state-of-the-art performance by training the encoder with contrastive loss, where given a job the model uses 1) random resumes as in-batch negatives and 2) rejected resumes as hard negatives, and vice versa for resumes. However, due to label scarcity (\Cref{sec:Background}), the quantity of rejected resumes/hard negatives is highly limited.


We propose \RunnerUpMining{} (\RunnerUpMiningShort{}), a new hard-negative mining method that selects “runner-up” resume-job pairs as hard negatives based on compatibility scores from an encoder model.
In dense text retrieval tasks, many prior hard-negative mining methods that treat \emph{unlabeled or incorrect top-k results} (e.g., using BM25) as hard \emph{negatives} \cite{xiong2021approximate,Zhao2024}.
However, in resume-job matching, we find many of these top-k results are \emph{positive} pairs that are unlabeled simply because the candidate did not apply to the job\footnote{For example, he/she may be overqualified or may have already accepted another offer.}.
To avoid these top-k false negatives, we instead take ``runner-up'' (e.g., top 3\%-4\%) samples based on the cosine similarity scores, and use them as hard negatives for training.


Specifically, \RunnerUpMiningShort{} begins by computing the cosine similarity scores between all possible resume-job pairs using an encoder model $f_\mu$ (e.g., \confitsimple{}).
Then, we rank these pairs by similarity score and randomly sample resumes and jobs from the top percentile \emph{ranges} (e.g., top 3\%-4\%) as challenging hard negatives.
Finally, during contrastive training, we replace hard negatives used in \confitold{} (i.e., rejected resumes/jobs) with these hard negatives mined by \RunnerUpMiningShort{}.


\subsection{\framework{}}
\label{subsec:Confit_overall}
We summarize our contribution in \framework{}. First, we simplified the encoder architecture used by \confitold{}, so that complex dynamics between different resumes and job posts are learned directly from the data (i.e., \confitsimple{}, \Cref{subsec:Model Architecture}).
Then, we improve the encoder training process by using 1) \HyReShort{} to generate pivot hypothetical resumes to simplify the job post representation space during both training and testing; and 2) \RunnerUpMiningShort{} to mine high-quality hard negatives for more effective contrastive training.
We illustrate \framework{}'s inference process in \Cref{fig:inference}.
Given a job post, \framework{} converts it into an embedding by first using an LLM to generate a hypothetical resume, and then using the encoder to produce a job embedding using the concatenation of the generated resume and the original job post. Given a resume, \framework{} directly uses the encoder to produce the resume embedding. The compatibility score between a resume and a job is then computed as the cosine similarity between their embeddings.

To train \framework{} encoder, we use \HyReShort{} to augment all job posts with generated reference resumes (\Cref{subsec:Hypothetical Resume Embedding}), and replace them with original job posts for later training and testing.
Then, we use \RunnerUpMiningShort{} to mine hard-negative resumes and (augmented) jobs (\Cref{subsec:Hard Negative Mining}).
Finally, we train the simplified \confitsimple{} using the modified contrastive learning loss $\mathcal{L}$ from \citet{confit_v1}:
\begin{align}
  &\qquad\qquad \mathcal{L} = \mathcal{L}_{R} + \mathcal{L}_{J} \label{eq:contrastive_loss} \\
  \mathcal{L}_{R} = &-\log \frac{e^{f_{\theta}(R_i^{+},J_i^{+})}}{e^{f_{\theta}(R_i^{+},J_i^{+})} + \sum_{j=1}^{l} e^{f_{\theta}(R_i^{+},J_{i,j}^{-})}} \nonumber \\
  \mathcal{L}_{J} = &-\log \frac{e^{f_{\theta}(R_i^{+},J_i^{+})}}{e^{f_{\theta}(R_i^{+},J_i^{+})} + \sum_{j=1}^{l} e^{f_{\theta}(R_{i,j}^{-},J_i^{+})}} \nonumber
\end{align}
where $\mathcal{L}_R$ and $\mathcal{L}_J$ is the resume/job contrastive loss, respectively; $(R^+, J^+)$ denotes accepted resume-job pairs; $(R^-, J^+)$ or $(R^+, J^-)$ denote negative resume-job pairs including both in-batch negatives and hard negatives; and $f_\theta$ is cosine similarity after embedding the resume-job pairs.


\begin{table*}[!t]
  \centering
  \scalebox{0.63}{
    \begin{tabular}{ll cccc cccc }
       \toprule
      & & \multicolumn{4}{c}{\textbf{Recruiting Dataset}} & \multicolumn{4}{c}{\textbf{AliYun Dataset}} \\
      & & \multicolumn{2}{c}{Rank Resume} & \multicolumn{2}{c}{Rank Job} 
      & \multicolumn{2}{c}{Rank Resume} & \multicolumn{2}{c}{Rank Job} \\
      \cmidrule(lr){3-4} \cmidrule(lr){5-6} \cmidrule(lr){7-8} 
      \cmidrule(lr){9-10}
      \textbf{Method} & \textbf{Encoder} 
      &  Recall@100 & nDCG@100 & Recall@10 & nDCG@10 & Recall@10 & nDCG@10 &  Recall@10 & nDCG@10   \\
      \midrule

      \multirow{2}{*}{RawEmbed.}
      & E5-base
      & 46.48 & 22.87 
      & 43.58 & 25.63

      & 49.52 & 29.66 
      & 36.24 & 27.19
      \\
      & text-embedding-3-large
       & 75.49 & 40.28 
      & 79.67 & 56.73  
      
      & 68.74 & 46.19 
      & 48.52 & 38.36 \\
      BM25
      & - 
      & 41.73 & 17.36
      & 38.92 & 24.14
      
      & 63.18 & 40.56 
      & 44.83 & 31.18
      \\
      \cmidrule(lr){2-10}
      
      MV-CoN
      & E5-base
      & 10.29 & 7.46 & 24.75 & 8.29
      
      & 11.01 & 6.47 & 14.63 & 8.29
      \\
      InEXIT
      & E5-base
      & 12.32 & 6.09 &19.33&  4.16 
      & 10.54 & 5.15 & 9.82 & 6.11 
    \\
     
      \confitold{}
      & E5-base
      & 65.13 & 28.74 & 68.42 & 40.85
      
      &  88.08 & 70.39 &  61.58& 43.45
       \\
     \cmidrule(lr){2-10}
      \framework{} (ours)
     
 
 & E5-base
 
       & \textbf{84.44} & \textbf{48.40}
       & \textbf{88.67}  & \textbf{66.61}
       & \textbf{96.18} & \textbf{82.20}
       & \textbf{82.11} & \textbf{67.96} \\
      \bottomrule
    \end{tabular}
  }
  \caption{Comparing ranking performance of various approaches with E5-base as backbone encoder. Results for non-deterministic methods are averaged over 3 runs. Best result is shown in \textbf{bold}.}
  \label{tbl:main_exp_bert}
  \vspace{-5pt}
\end{table*}
\begin{table*}[!t]
\centering
\scalebox{0.63}{
    \begin{tabular}{ll cccc cccc}
      \toprule
      & & \multicolumn{4}{c}{\textbf{Recruiting Dataset}} & \multicolumn{4}{c}{\textbf{AliYun Dataset}} \\
      & & \multicolumn{2}{c}{Rank Resume} & \multicolumn{2}{c}{Rank Job} 
      & \multicolumn{2}{c}{Rank Resume} & \multicolumn{2}{c}{Rank Job} \\
      \cmidrule(lr){3-4} \cmidrule(lr){5-6} \cmidrule(lr){7-8} 
      \cmidrule(lr){9-10}
      \textbf{Method} & \textbf{Encoder} 
      &  Recall@100 & nDCG@100 & Recall@10 & nDCG@10 & Recall@10 & nDCG@10 &  Recall@10 & nDCG@10   \\
      \midrule
      \multirow{2}{*}{RawEmbed.}
      & Jina-v2-base
       & 54.73 & 26.34  
      & 44.38 & 27.64 
      & 57.20 & 36.72 
      & 43.27 & 32.29 \\
      
      & text-embedding-3-large
       & 75.49 & 40.28 
      & 79.67 & 56.73  
      
      & 68.74 & 46.19 
      & 48.52 & 38.36 \\
      BM25
      & - 
      & 41.73 & 17.36
      & 38.92 & 24.14
      
      & 63.18 & 40.56 
      & 44.83 & 31.18
      \\
      \cmidrule(lr){2-10}
      MV-CoN
      & Jina-v2-base
      
       & 10.50 & 4.06
       &19.83 & 5.72
       & 9.11&4.72
       & 13.50 &9.99\\
      InEXIT
      & Jina-v2-base
       & 8.83 &4.01 
       &6.50& 3.05
       & 11.36 &4.83
       & 13.49& 10.46\\
      \confitold{}
      & Jina-v2-base
      
       & 71.28 & 34.79
       & 76.50 & 52.57
  
       & 87.81 &65.06 
       &72.39 &56.12\\
       
      \cmidrule(lr){2-10}
      \framework{} (ours)
      
      
       
   
    & Jina-v2-base 
      & \textbf{86.13}&\textbf{51.90} & \textbf{94.25}&\textbf{73.32}
      & \textbf{97.07} & \textbf{83.11} &\textbf{ 80.49} & \textbf{68.02}\\
      
      \bottomrule
    \end{tabular}
  }
  \caption{Comparing ranking performance of various approaches with Jina-v2-base as backbone encoder. Results for non-deterministic methods are averaged over 3 runs. Best result is shown in \textbf{bold}.}
  \vspace{-10pt}
\label{tbl:main_exp_e5}
\end{table*}

\section{Experiments}
\label{sec:Experiments}
We evaluate \framework{} on two real-world person-job fit datasets, and measure its performance on ranking resumes and ranking jobs.

\subsection{Dataset and Preprocessing}
\label{subsec:Dataset and Preprocessing}

\paragraph{AliYun Dataset} To our knowledge, the 2019 Alibaba job-resume intelligent matching competition\footnote{
\href{https://tianchi.aliyun.com/competition/entrance/231728/introduction}{https://tianchi.aliyun.com/competition/entrance/231728}
} \emph{provided} the only publicly available person-job fit dataset.
All resume and job posts were desensitized and were processed into a collection of text fields, such as ``Education'' and ``Work Experiences'' for a resume (see \Cref{sec:More Details on Dataset and Preprocessing} for more details). All resumes and jobs are in Chinese.

\paragraph{Recruiting Dataset} The resumes and job posts are provided by a hiring solution company. To protect the privacy of candidates, all records have been anonymized by removing sensitive identity information (see \Cref{sec:More Details on Dataset and Preprocessing}).
For each resume-job pair, we record whether the candidate is accepted ($y=1$) or rejected ($y=0$) for an interview.
Similar to the AliYun dataset, all resumes and jobs were available as a collection of sections/fields. Both English and Chinese resumes and jobs are included.

\subsection{Implementation Details}
\label{subsec:Implementation Details}
We follow \confitold{} \cite{confit_v1} and experiment with two encoder architectures. We consider E5-base \cite{E5}, and Jina-v2-base.
\cite{mohr2024multi}. Both encoders are trained on large-scale Chinese-English bilingual text data, and were amongst the best embedding models according to benchmarks such as MTEB \cite{muennighoff2022mteb} at the time of the project.
For \HyReShort{}, we use GPT-4o-mini as the LLM for generating hypothetical resumes due to its cost-effectiveness.
We provide the prompts used in \Cref{subsec:HyRe Prompts}.
We then train \confitsimple{} with \HyReShort{} using contrastive learning (without hard negatives), and use to mine hard negatives with \RunnerUpMiningShort{}.
For \RunnerUpMiningShort{}, we use the top 3\%-4\% percentile to pick hard-negatvies \emph{across all experiments}.
Finally, we train \confitsimple{} again, using 1) \HyReShort{} augmented job data; and 2) \RunnerUpMiningShort{} mined hard negatives. We use Eq \eqref{eq:contrastive_loss} with a batch size of 4 and 2 hard negatives per batch, and a learning rate of 1e-5.
This is the final encoder used for \framework{}.
We run all experiments using one A100 80GB GPU.
To speed up cosine similarity computation across a large pool of resumes and jobs, we use FAISS \cite{FAISS} throughout all experiments.



\subsection{Baselines}
\label{subsec:Baselines}
We compare \framework{} against both recent best person-job fit systems and strong baselines from information retrieval systems.

Prior person-job fit systems include \emph{MV-CoN} \cite{mvcon}, \emph{InEXIT} \cite{InEXIT}, and \emph{ConFit} \cite{confit_v1} and more \cite{APJFNN,pjfnn,DPGNN}.
MV-CoN considers a co-teaching network \cite{han2018coteaching} to learn from sparse, noisy person-job fit data, and InEXIT uses hierarchical attention to model interactions between the text fields of a resume-job pair.
\confitold{} focuses on dense retrieval techniques, and uses contrastive learning and LLM-based data augmentation to train an encoder.
Other older methods such as APJFNN \cite{APJFNN}, PJFNN \cite{pjfnn}, and DPGNN \cite{DPGNN} are omitted since they are already outperformed by these more recent methods.


We also compare against methods from information retrieval systems such as: \emph{BM25} \cite{bm25,bm25-all} and \emph{RawEmbed}. BM25 is a strong baseline used for many text ranking tasks \cite{thakur2021beir,kamalloo2023beir-resources}, and RawEmbed is based on dense retrieval methods \cite{DPR,FAISS} that directly uses a pre-trained encoder to produce dense embeddings for inner product scoring.
Since RawEmbed does not need training, we consider both open-source encoders such as E5 \cite{E5} and Jina-v2 \cite{mohr2024multi} and commercial models such as OpenAI's text-embedding-003-large \cite{text-ada}.



\begin{figure}[!t]
    \centering
    \includegraphics[scale=0.36]{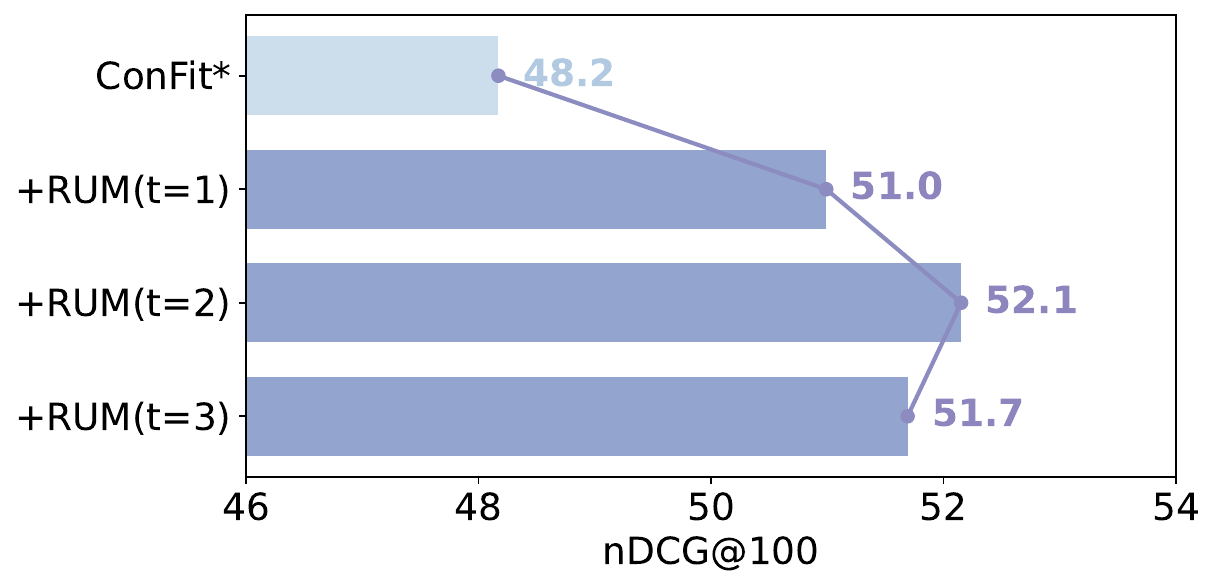}
    \caption{Iterative RUM using Jina-v2-base as encoder model. ``\RunnerUpMiningShort{}(t=N)'' indicates applying \RunnerUpMiningShort{} $N$ times.}
    \label{fig:iterative}
    \vspace{-18pt}
\end{figure}

     
       

\begin{table*}[!t]
\centering
\scalebox{0.75}{
    \begin{tabular}{l cccc cccc}
      \toprule
      & \multicolumn{4}{c}{\textbf{Recruiting Dataset}}
      & \multicolumn{4}{c}{\textbf{AliYun Dataset}}\\
      & \multicolumn{2}{c}{Rank Resume} & \multicolumn{2}{c}{Rank Job} 
      & \multicolumn{2}{c}{Rank Resume} & \multicolumn{2}{c}{Rank Job}\\
      \cmidrule(lr){2-3} \cmidrule(lr){4-5} 
      \cmidrule(lr){6-7} \cmidrule(lr){8-9} 
      \textbf{Method}
      & Recall@100 & nDCG@100 & Recall@10 & nDCG@10
      & Recall@100 & nDCG@100 & Recall@10 & nDCG@10\\
      \midrule
      \confitold{}
      & 71.28 & 34.79 & 76.50 & 52.57 
      & 87.81 & 65.06 & 72.39 & 56.12
      \\
      \cmidrule{2-9}
      \confitsimple{}
      & 82.53 & 48.17 & 85.58 & 64.91
      & 94.90&78.40 & 78.70& 65.45
       \\
      +\HyReShort{}
       &85.28 & 49.50
       &90.25 & 70.22
       & 96.62&81.99 & \textbf{81.16}& 67.63
       \\
      +\RunnerUpMiningShort{}
       & \textbf{86.13}&\textbf{51.90} & \textbf{94.25}&\textbf{73.32}
       & \textbf{97.07}&\textbf{83.11} & 80.49& \textbf{68.02}
       \\
      \bottomrule

    \end{tabular}
  }
\caption{\framework{} ablation studies. ``\confitsimple{}'' refers using a simplified encoder architecture. \framework{} trains \confitsimple{} with \RunnerUpMiningShort{} and \HyReShort{}. We use Jina-v2-base as the encoder due to its better performance.
}
\label{tbl:ablation}
\end{table*}
\vspace{-0.2cm}
\subsection{Main Results}
\label{subsec:Main Results}
\Cref{tbl:main_exp_bert} summarizes \framework{}'s performance in comparison to other baselines on the Recruiting and AliYun datasets when E5-base is used as the encoder.
We report two ranking metrics, Recall@K and nDCG@K.
In general, \confitold{} shows substantial improvement over many prior methods, such as MV-CoN, InEXIT, and BM25.
\framework{} further improves \confitold{} by 17.1\% in recall and 20.4\% in nDCG score on average, outperforming all other baselines, including OpenAI's text-embedding-003-large, on both datasets.

\Cref{tbl:main_exp_e5} summarizes each method's performance when a different encoder (Jina-v2-base) is used.
Similar to \Cref{tbl:main_exp_bert}, \confitold{} shows impressive performance compared to many baselines, but is exceeded by OpenAI's text-embedding-003-large.
However, \framework{} outperforms all other baselines, by up to 10.6\% in recall and 14.5\% in nDCG score for ranking resumes and jobs on both datasets\footnote{We believe this is due to, despite Jina-v2's recency, \citet{günther2023jina} find E5 to be more robust for retrieval tasks.}.
We believe these results underscore the effectiveness of our method across different encoders and datasets.



%
%

%
%

%

%
%
\begin{figure}[!t]
    \centering
    \includegraphics[scale=0.45]{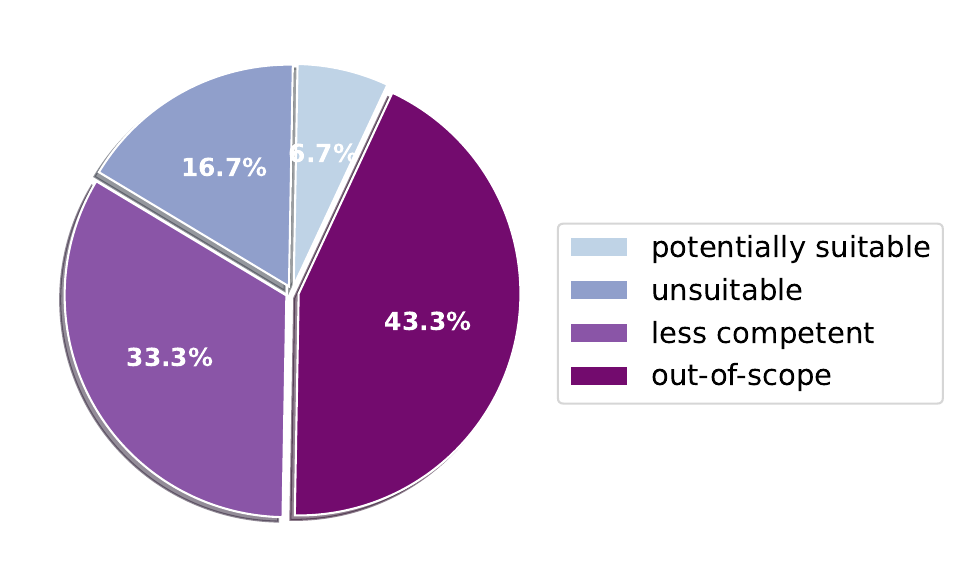}
    \caption{\framework{} error analysis. We find 43\% of the errors made are due to reasons not identifiable using resume/job documents alone, and 33\% due to a candidate’s resume satisfying all the job requirements but is less competent than other competing candidates}
    \label{fig:error_analysis}
\end{figure}

\subsection{Iterative \RunnerUpMiningShort{}}
Since \RunnerUpMiningShort{} only requires an encoder model to mine hard negatives, we also experiment with using \RunnerUpMiningShort{} iteratively to improve the model's performance.
For each iteration, we 1) use the encoder trained from the previous iteration to mine hard negatives using \RunnerUpMiningShort{}; and 2) train the backbone encoder again using the newly acquired hard negatives\footnote{
  We also experimented with continuously training the model from the previous iteration, but found that it can more easily lead to overfitting.
}.
Throughout all iterations, we keep all hyperparameters the same as in \Cref{subsec:Implementation Details}. We focus on the ranking resume task in the Recruiting dataset since it is the most challenging. 

We present the results in \Cref{fig:iterative}. In general, we find that training with more than one iteration ($t>1$) improves resume ranking performance compared to a single iteration ($t=1$). However, as the number of iterations increases, improvement begins to drop.
We believe this is similar to many model self-improvement research \cite{llm-can-self-improve,yu2024teachinglanguagemodelsselfimprove}. While model-created data often helps improve performance, these data also contains noises that can compound over multiple iterations.
We believe techniques such as model ensembles \cite{zhang2011robust,moreira2024nvretrieverimprovingtextembedding} could help mitigate this issue, and we leave this for future work.







\subsection{Ablation Studies}
\label{subsec:Ablation Studies}

\Cref{tbl:ablation} presents our ablation studies for each component of \framework{}.
First, we consider only simplifying the encoder architecture (denoted as \emph{ConFit$^{*}$}), and use rejected resumes/jobs as hard negatives used by \confitold{} \cite{confit_v1}.
We find that this simplification significantly improves ranking performance.
We believe this is due to the strong performance of recent embedding models, which can effectively learn complex interactions between resumes and job posts directly from data.
Next, we use \HyReShort{} to generate reference resumes, augment job posts, and train \confitsimple{} on the augmented data (denoted as \emph{+\HyReShort{}}).
We find this to improve 2.9\% absolute in recall and 3.1\% nDCG scores on average.
Finally, we use \RunnerUpMiningShort{} to mine hard resume/(augmented) job negatives (denoted as \emph{+RUM}), and replace them with the rejected resume/jobs used by \confitold{}.
In \Cref{tbl:ablation}, we find that \RunnerUpMiningShort{} further improved by, on average, 1.2\% absolute in recall scores, and 1.8\% absolute in nDCG across both datasets.
We find that all components are important for \framework{}.

For other results, such as comparing \RunnerUpMiningShort{} against methods that use BM25, with different percentile ranges other than the top 3\%-4\% as well as testing \RunnerUpMiningShort{} against hard negatives mined by BM25, please refer to \Cref{sec:Additional Results on RUM}.




\section{Analysis}
In this section, we provide both qualitative and quantitative analysis on the ranking output produced by \framework{}. We mainly focus on the Recruiting dataset as it is more challenging.

\subsection{Error Analysis}
To analyze the errors made by \framework{}, we manually inspect 30 \emph{negative} resume-job pairs from the ranking tasks that are \emph{incorrectly ranked at top 5\%} and are before at least one positive pair.
For each incorrectly ranked pair, we compare against other positive resume-job pairs from the dataset, and categorize the errors following the criteria from \citet{confit_v1}: \emph{unsuitable}, where some requirements in the job post are not satisfied by the resume; \emph{less competent}, where a resume satisfies all job requirements, but many {competing candidates} have a higher degree/more experience; \emph{out-of-scope}, where a resume satisfies all requirements, appears competitive compared to other candidates, but is still rejected due to other (e.g., subjective) reasons not presented in our resume/job data themselves; and \emph{potentially suitable}, where a resume from the ranking tasks satisfied the requirements and seemed competent, but had no label in the original dataset.

We present our analysis in \Cref{fig:error_analysis}. We find a significant portion of the errors are \emph{out-of-scope}, where it is hard to determine the reason for rejection based on information in the resume/job data alone.
The next most frequent error is \emph{less competent}, which is understandable since \framework{} produces a compatibility score for a resume-job pair independent of other candidates.
Lastly, we also find that 16.7\% of the errors are \emph{unsuitable}, with resumes not fulfilling certain requirements such as ``Backend SWEs with \emph{Ruby} experience''.
We believe \emph{unsuitable} errors can be mitigated by combining \framework{} with keyword-based approaches (e.g., BM25 or human-designed rules). We leave this for future work.

\begin{table}[!t]
    \centering
    \scalebox{0.8}{
    \begin{tabular}{lcccc}
    \toprule
    \textbf{Method} & \textbf{Gender}  & \textbf{Male(\%)} & \textbf{Female(\%)} \\
    \midrule
    
    \confitsimple{}
    & \xmark & 70.43 & 29.57
    \\
    
    \confitsimple{}
    & \cmark & 75.63 & 24.37
    \\
    
    \confitsimple{} + \RunnerUpMiningShort{}
    & \xmark & 66.67 & 33.33
    \\
    
    \confitsimple{} + \RunnerUpMiningShort{}
    & \cmark & 69.72 & 30.28
    \\
    
    \bottomrule
    \end{tabular}
    }
    \caption{
        Gender distribution in the top-10 ranked resume, averaged across all test job postings. ``Gender'' indicates whether gender information is used during training. ``Male/Female'' indicates the proportion of the top-10 resume that belongs to a male/female candidate.
    }
    \label{tbl:gender}
\end{table}
\begin{figure}[!t]
    \centering
    \includegraphics[scale=0.28]{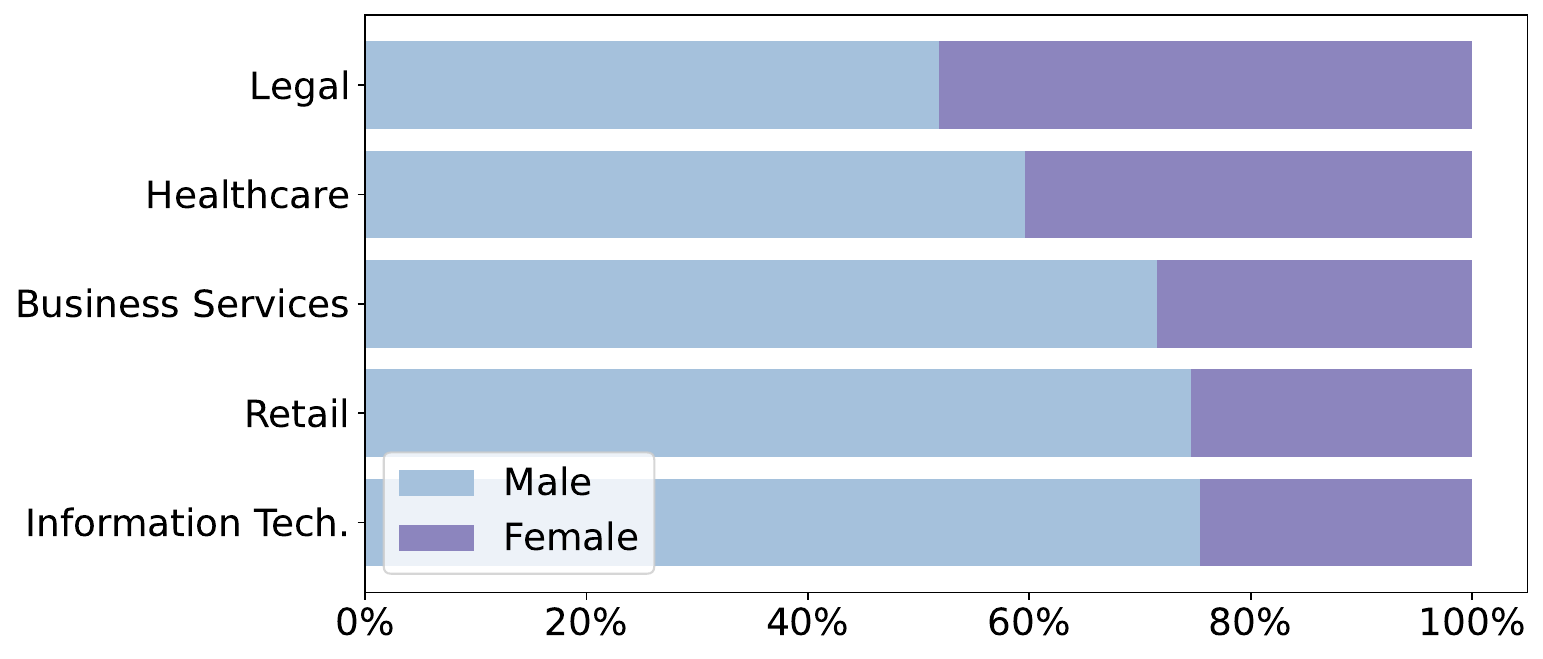}
    \caption{Gender distribution in different industries. We picked five industries to illustrate trend. In entire dataset, 72\% of the candidates are male, and 28\% are female.
    }
    \label{fig:dset_gender_dist}
\end{figure}

\subsection{Bias Analysis}
\label{subsec:bias analysis}
\framework{} relies on pretrained encoders such as E5 and Jina-v2 \cite{devlin2019bert,E5}, and it is well-known that many powerful encoders can contain biases \cite{pmlr-v97-brunet19a,social-bias,jentzsch-turan-2022-gender,Caliskan_2022} such as gender, age, demographics, etc.
As a case study, we examine whether ranking outputs from \framework{} contain gender bias.
For this experiment, we add gender information\footnote{Since gender information is voluntarily submitted by the candidates, this affects around 15\% of all resumes.} back to our desensitized resumes, and compare 1) whether training on gender-included data increases gender disparities in the ranking output; and 2) whether \RunnerUpMiningShort{} has any effect on gender bias.

We present our analysis in \Cref{tbl:gender}. Across all experiments, we find adding gender information back into the training data increases gender disparities in the ranking output. ``\confitsimple{}'' trained with gender information widened the gender gap in the ranking output by $\sim$10\% compared to training without gender, and ``\confitsimple{}+\RunnerUpMiningShort{}'' widened the gap by $\sim$6\%.
In general, we believe one major cause of this disparity is the inherent biases in the datasets.
In \Cref{fig:dset_gender_dist}, we find gender distributions are uneven across different industries, and that overall, 72\% of the (accepted or rejected) resumes used during training are from male candidates, and only 28\% are from female candidates.

In \Cref{tbl:gender}, we also find that using \RunnerUpMiningShort{} helps reduce gender disparities compared to \confitsimple{} alone.
We believe this indicates \RunnerUpMiningShort{} provided challenging hard negatives that do \emph{not} always rely on gender information, hence reducing gender disparity during training. In general, we advise researchers and practitioners to be cautious when using \framework{}, and to use explicit de-biasing techniques \cite{bolukbasi2016man,cheng2021fairfil,gaci-etal-2022-debiasing,guo-etal-2022-auto} \emph{in addition to} data de-sensitization used in this work. We do not condone the use of \framework{} for any ethically unjust purposes.

\section{Related Work}
\label{sec:Related Work}

\paragraph{Person-job fit systems}

Early person-job fit systems that use neural networks typically focus on architecture designs. These methods include \citet{APJFNN,pjfnn,cnn-lstm-siamese,jiang2020learning,10169716}, which explores
architectures such as RNN, LSTM \cite{staudemeyer2019understanding} and CNN \cite{oshea2015introduction}.
However, these are significantly outperformed by more recent transformer-based methods such as \citet{siamese,cnn-lstm-siamese,bian-etal-2019-domain,Zhang2023FedPJFFC}, which focuses on small architecture or loss modifications.
For example, MV-CoN \cite{mvcon} uses a co-teaching network \cite{malach2018decoupling} to perform gradient updates based on model's confidence to data noises; InEXIT \cite{InEXIT}, which uses hierarchical attention to model resume-job interactions; and DPGNN \cite{DPGNN}, which uses a graph-based approach with a novel BPR loss to optimize for resume/job ranking.
However, these methods tend to focus on task-specific modifications, such as assuming no unseen resumes/jobs at test time or assuming access to internal data such as whether the recruiter sent private messages to the job applicant.
More recently, \confitold{} \cite{confit_v1} focuses on applying dense retrieval techniques to person-job fit. Using contrastive learning with encoders such as E5 \cite{E5}, \citet{confit_v1} demonstrates its performance and flexibility by achieving the best scores in almost all ranking and classification tasks across two different person-job fit datasets.
In this work, we propose \HyReShort{} and \RunnerUpMiningShort{} to further enhance \framework{}, achieving an average absolute improvement of 13.8\% in recall and 17.5\% in nDCG across job ranking and resume-ranking tasks in two resume-job benchmarks.

\paragraph{Dense retrieval techniques} 
\framework{} benefits from dense retrieval techniques such as contrastive learning \cite{SimCLR,radford2021learning} and hard-negative mining.
Popular methods in text retrieval include BM25 \cite{bm25,bm25-all}, a keyword-based approach used as the baseline in many text ranking tasks \cite{ms-marco,thakur2021beir,muennighoff2022mteb}, and dense retrieval methods such as \citet{DPR,izacard2021contriever,E5,günther2023jina}, which uses contrastive learning with an encoder to obtain high-quality passage embeddings, and typically performs top-k search based on inner product.

To further improve retrieval results, researchers considered methods such as query/document expansions \cite{query-expansion-survey} and hard-negative mining \cite{robinson2021contrastive}.
Recent query/document expansion approaches include HyDE \cite{hyde} and HyQE \cite{hyqe}, which prompts an LLM to augment the original query or the passage to retrieve and is typically used \emph{without} finetuning (the LLM or the encoder model). Common hard-negative mining strategies often use BM25 to select top-k unlabeled/incorrect samples as hard negatives \cite{DPR,Zhao2024,nguyen-etal-2023-passage}. Other methods include: ANCE \cite{xiong2021approximate} which samples negatives from the top retrieved documents asynchronously during training; SimANS \cite{zhou-etal-2022-simans} which samples hard negatives using a manually designed probability distribution; and more \cite{zhan2021,syneg}. \confitold{} \cite{confit_v1} presents the first successful attempt to use contrastive learning for person-job fit.
We extend \confitold{} and adapt recent dense retrieval techniques to person-job fit by 1) sampling hard-negative resumes/jobs from a ``runner-up'' \emph{percentile ranges} to mitigate selecting false-negatives; and 2) few-shot prompting an LLM to augment a job posted used for later encoder \emph{training}.

\section{Conclusion}
\label{sec:Conclusion}
We propose \framework{}, an improvement of \confitold{} to model person-job fit. Similar to \confitold{}, we model person-job fit using dense retrieval techniques such as contrastive learning. Unlike \confitold{}, we first simplified the encoder architecture, and improved encoder training using 1) hypothetical reference resume generated by an LLM to augment a job post; and 2) a new hard-negative mining strategy that selects ``runner-up'' resume-job pairs to avoid false negatives used for encoder training.
We evaluate \framework{} on two real-world datasets, and demonstrate that it outperforms \confitold{} and prior methods, achieving an average \emph{absolute} improvement of 13.8\% in recall and 17.5\% in nDCG across both ranking resume and ranking job tasks.
We believe our work lays a strong foundation for future resume-job matching systems to leverage the latest advancements in dense text retrieval.

\section{Limitations}

\paragraph{Data Sensitivity} To our knowledge, there is no standardized, public person-job fit dataset\footnote{The AliYun dataset used in this work is no longer publicly available as of 09-11-2023.} that can be used to compare performances of existing systems \cite{pjfnn,APJFNN,mvcon,DPGNN,InEXIT}.
This is due to the highly sensitive nature of resume and job post content, making large-scale person-job fit datasets largely proprietary.
We follow \citet{confit_v1} and provide the best effort to make \framework{} reproducible and extensible for future work.
We will open-source implementations of \framework{}, related baselines, data processing scripts, and dummy train/valid/test data files that can be used to test drive our system end-to-end.  We will also privately release our model weights and full datasets to researchers under appropriate license agreements, aiming to make future research in person-job fit more accessible.

\paragraph{Local \HyReShort{}} 

During \HyReShort{}, we use a commercial LLM such as GPT-4o-mini to generate hypothetical reference resumes to augment our job data for retrieval.
Ideally, one would prefer using an in-house, local model to better retrain data privacy and optimize for inference speed.
However, in our prior study, we find fine-tuning open-source LLMs (up to 32B in size) for this task to be challenging, due to the scarce and subjective nature of the person-job fit labels (\Cref{subsec:Finetuning HyRe}).
We believe designing data/training algorithms tailored for this task could benefit person-job fit systems such as \framework{}, which we leave for future work.


\section{Ethical Considerations}

\framework{} uses and trains pretrained encoders such as E5 \cite{E5} and Jina-v2 \cite{günther2023jina}, and it is well-known that these models contain biases including but not limited to gender, race, demographics, and more \cite{pmlr-v97-brunet19a,social-bias,jentzsch-turan-2022-gender,Caliskan_2022}.
For practical person-job fit systems, we believe it is crucial to ensure that they do \textbf{not} exhibit these biases, such as preferring a certain gender for certain jobs.
In this work, we have followed best practices from prior work \cite{DPGNN,confit_v1} to remove sensitive information (e.g., gender) from our data.
However, our bias analysis (\Cref{subsec:bias analysis}) reveals that gender disparities in the trained model could stem from imbalances in the dataset.
We do \textbf{not} condone using \framework{} for real-world applications without using de-biasing techniques such as \citet{bolukbasi2016man,cheng2021fairfil,gaci-etal-2022-debiasing,guo-etal-2022-auto,schick-etal-2021-self}, and in general, we do \textbf{not} condone the use of \framework{} for any morally unjust purposes.
To our knowledge, there is little work on investigating or mitigating biases in existing person-job fit systems, and we believe this is a crucial direction for future person-job research.

\bibliography{custom}

\setcounter{table}{0}
\renewcommand{\thetable}{A\arabic{table}}
\setcounter{figure}{0}
\renewcommand{\thefigure}{A\arabic{figure}}
\appendix
\clearpage

\section{More Details on Model Architecture}
\label{sec:more_details_on_model_architecture}

\begin{figure*}[!h]
    \centering
    \includegraphics[scale=0.75]{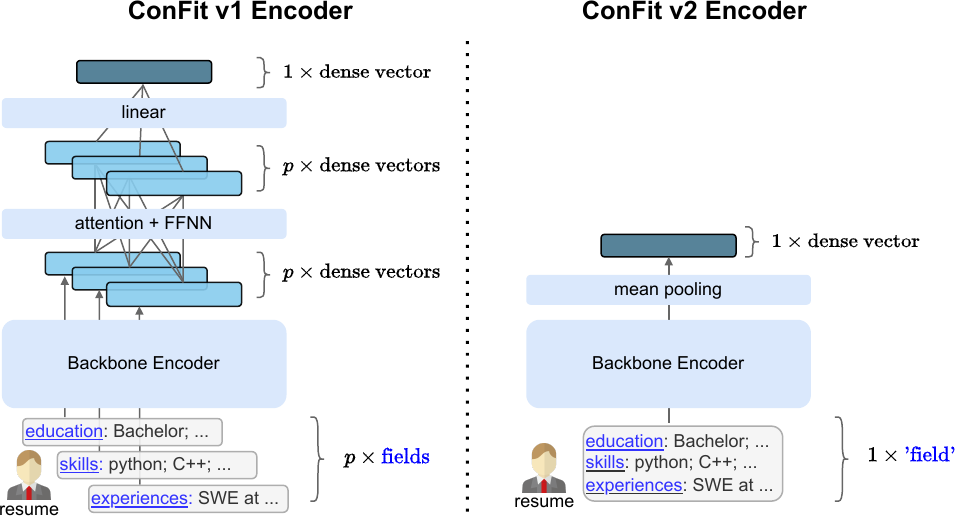}
    \caption{Encoder architecture comparison between \confitold{} (left) and \framework{} (right).}
    \label{fig:archi_diff}
\end{figure*}

\Cref{fig:archi_diff} displays encoder model architecture used in \confitold{} and \confitsimple{}. We remove all the additional layers used after the backbone encoder, including the attention layers that interact among features of each field, and the MLP layers for feature fusion. Instead, we directly employ mean pooling on the embeddings of all the tokens in the texts to obtain document-level representation \cite{E5,günther2023jina}.
Then, we perform contrastive learning on the document-level embeddings.

We implement this with both encoders such as E5 \cite{E5} and long-context encoders such as Jina-v2 \cite{günther2023jina}. E5 supports 512 input sequence length and Jina-v2 supports 8192. We add $L_2$ normalization to the features, and use temperature $T$ to rescale the cosine similarity to facilitate training. For E5 we use $T=0.01$, and for Jina-v2 we use $T=0.05$. In \Cref{tbl:main_exp_bert} and \Cref{tbl:main_exp_e5}, we find that \confitsimple{} significantly improve upon \confitold{} across all settings.



\section{Additional Results on \HyReShort{}}
\label{sec:Additional Results on HyRe}

\subsection{\HyReShort{} Prompts}
\label{subsec:HyRe Prompts}

We use the prompt in Table \ref{tbl:prompt} to perform few-shot prompting for hypothetical resume generation. In the prompt, \textit{\{\{\{example job\}\}\}}, \textit{\{\{\{example resume\}\}\}} are randomly sampled positive pairs from the training set, and the \textit{\{\{\{target job\}\}\}} is the target job to be augmented. For the ablation study in \ref{subsec: Hypothetical job}, we swap the order of template job and resume, send in a resume as target, and revise the prompt correspondingly. When concatenating job and hypothetical resume, we add \textit{["An Example Resume"]} in between for clarity.

\subsection{\HyReShort{} Upperbound}
\label{subsec:HyRe Upperbound}
In this section, we estimate the upper bound performance of \HyReShort{} by using real resumes from the dataset to augment each job during both training and testing.
However, evaluation based on training the entire dataset is time-consuming, and thus we trained on a randomly sampled subset containing 20\% of the training data while keeping all other settings identical to those used for training on the full dataset.


We experiment with three strategies of selecting ``ideal'' reference resumes for each job from the dataset.
For \emph{max-match (human)}, we select the highest-scoring resume for each job according to additional annotations provided by a hiring solution company. 
For \emph{max-match (Jina)}, we select the highest-scoring resume that achieves high human annotation score \textbf{and} a high cosine similarity score computed directly by Jina-v2 \cite{günther2023jina}.
For \emph{centroid}, we select the centroid resume of all accepted resumes given a job post.
Since computing the centroid resume across the full resume pool is time-consuming, we instead restrict the pool to all \emph{labeled} resumes given the job post.
Then, we pick the resume, which has the shortest average $L_2$ normalized distance (i.e. cosine similarity) to its job's accepted resumes, as the centroid resume.

We present the results in Table \ref{tbl:Upperbound}. Compared to \confitsimple{}, all three strategies show that \HyReShort{} yields a high potential performance if the hypothetical resumes can truly approximate the real selected ones.
This indicates that \HyReShort{}, if implemented well, could significantly improve ranking job or ranking resume performance.

\begin{table}[!t]
\centering
\scalebox{0.6}{
    \begin{tabular}{l cccc}
      \toprule
      & \multicolumn{4}{c}{\textbf{Recruiting Dataset}}\\
      & \multicolumn{2}{c}{Rank Resume} & \multicolumn{2}{c}{Rank Job} \\
      \cmidrule(lr){2-3} \cmidrule(lr){4-5} 
      \textbf{Method}
      &  Recall@100 & nDCG@100 & Recall@10 & nDCG@10 \\
      \midrule
      \confitsimple
       & 79.00 &44.21 &80.83&60.67
      \\
      \midrule
      Max-match (human)
      & 82.06 &73.14 &84.75 &70.11
      \\
      Max-match (Jina)
      & 80.14 &65.95 &82.78 &66.34\\

      Centroid
      & 74.82 &68.28 &87.50&69.29\\
    
      \bottomrule

    \end{tabular}
  }
\caption{Estimating \HyReShort{} upperbound using real resumes. To speed-up evaluation, we trained all models on a fixed subset of the data.}
\label{tbl:Upperbound}
\end{table}
\begin{table}[!t]
\centering
\scalebox{0.6}{
    \begin{tabular}{l cccc}
      \toprule
      & \multicolumn{4}{c}{\textbf{Recruiting Dataset}}\\
      & \multicolumn{2}{c}{Rank Resume} & \multicolumn{2}{c}{Rank Job} \\
      \cmidrule(lr){2-3} \cmidrule(lr){4-5} 
      \textbf{Method}
      &  Recall@100 & nDCG@100 & Recall@10 & nDCG@10 \\
      \midrule
      \confitsimple
       & 79.00 &44.21 &80.83&60.67
      \\
      \midrule
      Job2Resume
      & 80.92 &46.01 &85.00 &65.12
      \\
      Resume2Job
      
      & 74.18 &42.30 &77.75&59.82\\
    
      \bottomrule

    \end{tabular}
  }
\caption{Comparison of generating hypothetical job posts instead of resume.}
\label{tbl:Upperbound_job}
\end{table}
\begin{table}[!t]
\centering
\scalebox{0.55}{
    \begin{tabular}{l cccc}
      \toprule
      & \multicolumn{4}{c}{\textbf{Recruiting Dataset}}\\
      & \multicolumn{2}{c}{Rank Resume} & \multicolumn{2}{c}{Rank Job} \\
      \cmidrule(lr){2-3} \cmidrule(lr){4-5} 
      \textbf{Method}
      &  Recall@100 & nDCG@100 & Recall@10 & nDCG@10 \\
      \midrule
      \confitsimple
       & 79.00 &44.21 &80.83&60.67
      \\
      \midrule
      GPT-4o-mini (Few-shot)
      & 80.92 &46.01 &85.00 &65.12
      \\
     Qwen (SFT)
      
      & 71.62 & 42.07&78.50&57.38\\
      Qwen (SFT+DPO)
      
      &77.17 & 45.07&84.75&63.53\\
    
      \bottomrule

    \end{tabular}
  }
\caption{Comparison of Few-shot Prompting and Finetuning \HyReShort{}\ trained on subset}
\label{tbl:finetune}
\end{table}
\subsection{Hypothetical Job Embedding}
\label{subsec: Hypothetical job}

In \framework{}, we augment job data with hypothetical resumes generated by an LLM (\Cref{subsec:Hypothetical Resume Embedding}). Alternatively, one can also augment resume data with hypothetical jobs generated by an LLM, which we experiment in this section.

We follow the same setup as \Cref{subsec:HyRe Upperbound}, but instead generate a hypothetical job for each resume. We then compare this against generating hypothetical resumes (\HyReShort{}). We present the results in \Cref{tbl:Upperbound_job}.
We find, on average, augmenting job post with hypothetical resume is more effective.
We believe this is because job data are often short and contain less details compared to resumes, making it more beneficial for content expansion.

\subsection{Finetuning HyRe}
\label{subsec:Finetuning HyRe}

In this section, we examine whether fine-tuned LLMs are capable of generating hypothetical resume to approxiate \Cref{subsec:HyRe Upperbound}. We perform Supervised Fine-tuning (SFT) and Direct Preference Optimization (DPO) on Qwen-2.5 series models \cite{qwen2.5}. Specifically, we train Qwen-2.5-32B-Instruct with LoRA \cite{hu2022lora} on the Recruiting Dataset. All the experiments are implemented on 4 A100 80GB GPUs.

For SFT training, we select 6k high-quality resume-job pairs that have both a high human label score and a high cosine similarity score computed by Jina-v2' raw embeddings.
For DPO training, we select the resume with the highest human label score as the chosen sample, and randomly select a resume with lower label score as the rejected sample. This results in a ``preference'' dataset with around 13k pairs. We then train Qwen-2.5-32B-Instruct with SFT and DPO, and evaluate the checkpoints after each training stage.


To speed up evaluation, we follow \Cref{subsec:HyRe Upperbound} and evaluate all methods using a subset of the Recruiting dataset. We present the result in \Cref{tbl:finetune}.
We find after SFT+DPO training. the Qwen model improves \confitsimple{}, but it does not overpass Few-shot prompting on GPT-4o-mini.
We believe this is because accepting a resume is subjective, making it hard to learn an ``ideal candidate'' directly from the dataset itself.

\begin{table}[!t]
\centering
\scalebox{0.65}{
    \begin{tabular}{l cccc}
      \toprule
      & \multicolumn{4}{c}{\textbf{Recruiting Dataset}}\\
      & \multicolumn{2}{c}{Rank Resume} & \multicolumn{2}{c}{Rank Job} \\
      \cmidrule(lr){2-3} \cmidrule(lr){4-5} 
      \textbf{Method}
      &  Recall@100 & nDCG@100 & Recall@10 & nDCG@10 \\
      \midrule
      BM25(top-10)
      & 82.95 &44.62 &86.75 &64.36
      \\
      \midrule
      \RunnerUpMiningShort{}(0\%-1\%)
      & 80.80 &48.41 &86.00 &66.17 
      \\
      \RunnerUpMiningShort{}(1\%-2\%)
      & 85.13 &49.92 &85.13 &66.85
      \\
      \RunnerUpMiningShort{}(2\%-3\%)
      & 83.08 &49.09 &90.25 &70.21 
      \\
      \RunnerUpMiningShort{}(3\%-4\%)
      & 85.43 &\textbf{50.99} &\textbf{91.38} &\textbf{71.34} 
      \\
      \RunnerUpMiningShort{}(4\%-5\%)
      & \textbf{85.56}  &50.88 &87.25 &69.28
      \\
      \bottomrule

    \end{tabular}
  }
\caption{Comparing \RunnerUpMiningShort{} with BM25 methods, as well as \RunnerUpMiningShort{} under different percentiles ranges. We denote this as ``\emph{\RunnerUpMiningShort{}(L\%-H\%)}''.}
\label{tbl:RUM}
\end{table}
\begin{table}[!t]
  \centering
  \scalebox{0.75}{
    \begin{tabular}{l cc}
      \toprule
      Train & \textbf{Recruiting Dataset} & \textbf{Aliyun Dataset}\\
      \midrule
      \# Jobs     & 9469 & 19542 \\
      \# Resumes  & 41279 & 2718 \\
      \# Labels   & 48096 & 22124 \\
      \phantom{--}\textcolor{gray}{(\# accept)} & 19042 & 10185 \\
      \phantom{--}\textcolor{gray}{(\# reject)} & 29054 & 11939 \\
      \cmidrule(lr){1-3}
      \# Token per $R$ & 1303.5($\pm 947.3$) & 250.6($\pm 96.5$) \\
      \# Token per $J$ & 639.4($\pm 1548.7$) & 335.1($\pm 143.9$) \\
      \bottomrule
    \end{tabular}
  }
  \caption{Training dataset statistics. ``\# Token per $R$/$J$'' represent the \emph{mean($\pm$std)} number of token per resume/job after post-processing.}
  \label{tbl:train_dset}
\end{table}
%
%
\begin{table}[!t]
  \centering
  \scalebox{0.76}{
    \begin{tabular}{l ccc ccc}
      \toprule
      & \multicolumn{2}{c}{\textbf{Recruiting Dataset}} & \multicolumn{2}{c}{\textbf{AliYun Dataset}}\\
      \textbf{Test} & Rank $R$ & Rank $J$ & Rank $R$ & Rank $J$ \\
      \midrule
      \# Samples
      & 200 & 200
      & 300 & 300\\
      \# Jobs
      & 200 & 300
      & 300 & 2903\\
      \# Resumes
      & 3000 & 200
      & 1006 & 300\\
      \bottomrule
    \end{tabular}
  }
  \caption{Test dataset statistics.}
  \label{tbl:test_dset}
\end{table}

\begin{table*}[!t]
  \centering
  \scalebox{0.7}{
    \begin{tabular}{ll cc cc}
      \toprule
      & & \multicolumn{4}{c}{\textbf{Recruiting Dataset}} \\
      & & \multicolumn{2}{c}{Rank Resume} & \multicolumn{2}{c}{Rank Job} \\
      \cmidrule(lr){3-4} \cmidrule(lr){5-6}
      \textbf{Method} & \textbf{Encoder} 
      & nDCG@10-rand & nDCG@10-hard & nDCG@10-rand & nDCG@10-hard \\
      \midrule
      \multirow{3}{*}{RawEmbed.}
      & E5-base
      & 21.56 & 15.78 & 39.02 & 29.06\\
      & Jina-base
      & 11.36 & 12.42 & 34.28 & 34.30\\
      & text-embedding-003-large
      & 39.05 & 30.43 & 72.02 & 60.89\\
      \cmidrule(lr){2-6}
      BM25
      & - 
      & 27.01 & 15.47 & 44.77 & 33.76 \\
      \confitold{}
      & Jina-base
      & 33.65 & 24.57 & 65.67 & 54.67 \\
      \cmidrule(lr){2-6}
      \confitsimple{} + \RunnerUpMiningShort{} (ours)
      & Jina-base
      & 49.96 & 39.92
      & 84.05 & 73.69 \\
      \bottomrule
    \end{tabular}
  }
  \caption{Comparing ranking performance of various approaches when test set consists of hard samples mined by BM25. We denote test sets using rejected and random unlabeled samples as negatives \cite{confit_v1} as ``\emph{-rand}'', and test sets using rejected and unlabeled samples mined by BM25 as negatives as ``\emph{-hard}''.
  Since test sets are smaller, we use nDCG@10 for all cases.
  }
  \label{tbl:exp_hard_neg}
\end{table*}

\section{Additional Results on \RunnerUpMiningShort{}}
\label{sec:Additional Results on RUM}

We investigate two aspects of \RunnerUpMiningShort{}: (1) how different percentile ranges affect performance, and (2) how RUM compares to previous BM25-based hard-negative mining method \cite{DPR,Zhao2024}. Specifically, we consider “BM25(top-10)”, which retrieves top-10 highest-scoring unlabeled or incorrect resume/job computed by BM25.


We present the result in \Cref{tbl:RUM}.
Overall, we find \RunnerUpMiningShort{} consistently surpasses BM25-based negatives in nDCG across both ranking tasks. 
When using \RunnerUpMiningShort{}, we find although ranges such as 0\%–1\% outperforms BM25 in some metrics, it significantly underperforms other ranges such as 2\%-3\%, and 3\%-4\%.
This indicates that many high-scoring unlabeled samples are likely \emph{positives} for person-job fit.
In this work, we picked 3\%-4\% due to its high average performance, and \emph{fixed it for all subsequent runs} (e.g., with different model architectures and datasets).


\section{Details on Dataset and Preprocessing}
\label{sec:More Details on Dataset and Preprocessing}

\paragraph{Recruiting Dataset} The talent-job pairs are provided by a hiring solution company. The original resumes/job posts are parsed into text fields using techniques such as OCR. Some of the information is further corrected by humans. All sensitive information, such as names, contacts, college names, and company names, has been either removed or converted into numeric IDs. Example resume and job post are shown in \Cref{tbl:example_resume} and \Cref{tbl:example_job}, respectively.

\paragraph{AliYun Dataset} The 2019 Alibaba job-resume intelligent matching competition provided resume-job data that is already desensitized and parsed into a collection of text fields. There are 12 fields in a resume (\Cref{tbl:example_resume}) and 11 fields in a job post (\Cref{tbl:example_job}) used during training/validation/testing. Sensitive fields such as ``\chinese{居住城市}'' (living city) were already converted into numeric IDs. ``\chinese{工作经验}'' (work experience) was processed into a list of keywords. Overall, the average length of a resume or a job post in the AliYun dataset is much shorter than that of the Recruiting dataset (see \Cref{tbl:train_dset}).

We present the training and test dataset statistics in \Cref{tbl:train_dset} and \Cref{tbl:test_dset}, respectively.
\begin{table*}[!t]
  \centering
  \scalebox{1.0}{
    \begin{tabular}{p{0.95\linewidth}}
      \toprule
      \textbf{\HyReShort{} Prompt Template} \\
      \midrule
      Here is a template pair of matching resume and job:

[The start of the example job]

\textit{\{\{\{example job\}\}\}}

[The end of the example job]

[The start of the example resume]

\textit{\{\{\{example resume\}\}\}}

[The end of the example resume]

You are a helpful assistant. Following the above example pair of job and resume, construct an ideal 
resume for the target job shown below. You should strictly follow the format of the given pairs, make sure the resume you give perfectly matches the 
target job, and directly return your answer in plain text.

[The start of the target job]

\textit{\{\{\{target job\}\}\}}

[The end of the target job]\\

      \bottomrule
    \end{tabular}
  }
  \caption{Few-shot prompt template for hypothetical reference resume generation. ```\textit{\{\{\{...\}\}\}}''' are placeholders to be programmatically inserted during inference. ``target job'' is the job post to be augmented.
  }
  \label{tbl:prompt}
\end{table*}
\begin{table*}[h]
  \centering
  \scalebox{0.85}{
    \begin{tabular}{p{0.6\linewidth} p{0.45\linewidth}}
      \toprule
      \multicolumn{1}{c}{\textbf{$R$ from Recruiting Dataset}} &
      \multicolumn{1}{c}{\textbf{$R$ from AliYun Dataset}} \\
      \midrule
      \#\# languages
      & \#\# \chinese{期望工作城市}\\
      MANDARIN, ENGLISH
      & 551,-,-\\
      \#\# industries
      & \#\# \chinese{学历}\\
      IT.Electronic Industry
      & \chinese{大专}\\
      \#\# job functions
      & \#\# \chinese{期望工作行业}\\
      SOFTWARE\_AND\_MATHEMATICS
      & \chinese{房地产/建筑/建材/工程}\\
      \#\# experiences
      & \#\# \chinese{期望工作类型}\\
      most recent: Software Engineer for 8 years...
      & \chinese{工程造价/预结算}\\
      second most recent: ...
      & \#\# \chinese{当前工作行业}\\
      & \chinese{房地产/建筑/建材/工程}\\
      \bottomrule
    \end{tabular}
  }
  \caption{Example resume from the Recruiting dataset and AliYun dataset. The Recruiting dataset contains resumes in both English and Chinese, while the AliYun dataset contains resumes only in Chinese. All documents are available as a collection of text fields, and are formatted into a single string as shown above for \confitsimple{} training.}
  \label{tbl:example_resume}
\end{table*}
\begin{table*}[h]
    \centering
    \scalebox{0.85}{
      \begin{tabular}{p{0.6\linewidth} p{0.45\linewidth}}
        \toprule
        \multicolumn{1}{c}{\textbf{$J$ from Recruiting Dataset}} &
        \multicolumn{1}{c}{\textbf{$J$ from AliYun Dataset}} \\
        \midrule
        \#\# Basic Info
        & \#\# \chinese{工作名称}\\
        title: Sr. SWE-Perception Infra
        & \chinese{工程预算}\\
        job type: Full-Time
        & \#\# \chinese{工作城市}\\
        openings: 1
        & 719\\
        \#\# job functions
        & \#\# \chinese{工作类型}\\
        SOFTWARE\_AND\_MATHEMATICS
        & \chinese{工程造价/预结算}\\
        \#\# Requirements
        & \#\# \chinese{最低学历}\\
        Strong programming skills in C++
        & \chinese{大专}\\
        Minimum of a Masters Degree in CS or equivalent
        & \#\# \chinese{工作描述}\\
        2+ years of experience in software industry...
        & \chinese{能够独立完成土建专业施工预算...}\\
        \bottomrule
      \end{tabular}
    }
    \caption{Example job post from the Recruiting dataset and AliYun dataset. The Recruiting dataset contains job posts in both English and Chinese, while the AliYun dataset contains job posts only in Chinese. All documents are available as a collection of text fields, and are formatted into a single string as shown above for \confitsimple{} training.}
    \label{tbl:example_job}
\end{table*}

\end{document}